%% file: root.tex
\documentclass{salesforce_ai_research}
\usepackage[utf8x]{inputenc}
\usepackage[T1]{fontenc}
\usepackage{algorithm}
\usepackage{algorithmic}
\usepackage{amsmath,amssymb,mathtools}
\usepackage[table]{xcolor} 
\usepackage{tabularx}
\usepackage{booktabs}

\usepackage{array}
\usepackage{algorithm}
\usepackage{tikz}
\usepackage{pgfplots}
\pgfplotsset{compat=1.18}
\usetikzlibrary{arrows.meta,positioning,fit,calc,shapes,backgrounds,decorations.pathreplacing}
\usepackage{setspace}
\definecolor{myblue}{HTML}{4eb1fb}
\definecolor{myred}{HTML}{e56726}
\usepackage{graphicx} 
\usepackage{subcaption}
\usepackage{bm}
\usepackage{enumitem} 
\usepackage{listings}
\definecolor{codegreen}{rgb}{0,0.6,0}
\definecolor{codegray}{rgb}{0.5,0.5,0.5}
\definecolor{codepurple}{rgb}{0.58,0,0.82}
\definecolor{backcolour}{rgb}{0.95,0.95,0.92}

\lstdefinestyle{mystyle}{
    backgroundcolor=\color{backcolour},   
    commentstyle=\color{codegreen},
    keywordstyle=\color{magenta},
    numberstyle=\tiny\color{codegray},
    stringstyle=\color{codepurple},
    basicstyle=\ttfamily\scriptsize,
    belowcaptionskip=1\baselineskip,
    breakatwhitespace=false,         
    breaklines=true,                 
    keepspaces=true,                 
    numbers=left,       
    numbersep=5pt,                  
    showspaces=false,                
    showstringspaces=false,
    showtabs=false,                  
    tabsize=2,
    columns=fullflexible,
}

\lstdefinelanguage{yaml}{
  keywords={true,false,null},
  sensitive=false,
  comment=[l]{\#},
  morestring=[b]',
  morestring=[b]",
  moredelim=[l][\color{magenta}]{-\ },
  literate=
    {workflow_id:}{{{\color{blue!70!black}workflow\_id:}}}{12}
    {workflow_name:}{{{\color{blue!70!black}workflow\_name:}}}{14}
    {workflow_title:}{{{\color{blue!70!black}workflow\_title:}}}{15}
    {description:}{{{\color{blue!70!black}description:}}}{12}
    {running_config:}{{{\color{blue!70!black}running\_config:}}}{15}
    {variable_values:}{{{\color{blue!70!black}variable\_values:}}}{16}
    {injected_steps:}{{{\color{blue!70!black}injected\_steps:}}}{15}
    {category:}{{{\color{blue!70!black}category:}}}{9}
    {steps:}{{{\color{blue!70!black}steps:}}}{6}
    {step_number:}{{{\color{blue!70!black}step\_number:}}}{12}
    {Action:}{{{\color{blue!70!black}Action:}}}{7}
    {id:}{{{\color{blue!70!black}id:}}}{3}
}

\lstdefinelanguage{json}{
  keywords={true,false,null},
  sensitive=false,
  morestring=[b]",
  morecomment=[l]{/*},
  morecomment=[s]{/*}{*/},
  literate=
    *{:}{{{\color{blue!70!black}:}}}{1}
    {[}{{{\color{codegray}[}}}{1}
    {]}{{{\color{codegray}]}}}{1}
    {\{}{{{\color{codegray}\{}}}{1}
    {\}}{{{\color{codegray}\}}}}{1},
}

\usepackage{parskip}

\usepackage[all]{nowidow} 
\usepackage[protrusion=true,expansion=true]{microtype} 

\usepackage{lipsum} 
\usepackage{setspace}
\usepackage{multicol}
\usepackage{wrapfig,lipsum}
\setitemize{noitemsep,topsep=0pt,parsep=0pt,partopsep=0pt,leftmargin=8mm}
\newcommand*\samethanks[1][\value{footnote}]{\footnotemark[#1]}

\title{GPA: Learning GUI Process Automation from Demonstrations}
\author{Zirui Zhao\thanks{Correspondence to \texttt{\{zirui.zhao, junnan.li\}@salesforce.com}. GPA is a proof of concept project for research. For collaboration, product development, or enterprise application, feel free to contact us.} \quad Jun Hao Liew \quad Yan Yang \quad Wenzhuo Yang \quad Ziyang Luo \\\textbf{\quad Doyen Sahoo \quad Silvio Savarese \quad Junnan Li\samethanks}\\
\textit{Salesforce AI Research}\\\url{https://www.salesforceairesearch.com/gpa}
}

\begin{document} 
\maketitle

\begin{abstract}
    \underline{G}UI \underline{P}rocess \underline{A}utomation (GPA) is a lightweight but general vision-based Robotic Process Automation (RPA), which enables fast and stable process replay with only a single demo. Addressing the fragility of traditional RPA and the non-deterministic risks of current vision language model-based GUI agents, GPA introduces three core benefits: (1) \textbf{Robustness} via Sequential Monte Carlo-based localization to handle rescaling and detection uncertainty; (2) \textbf{Deterministic \& Reliability} safeguarded by readiness calibration; and (3) \textbf{Privacy} through fast, fully local execution. This approach delivers the adaptability, robustness, and security required for enterprise workflows. It can also be used as an MCP/CLI tool by other agents with coding capabilities so that the agent only reasons and orchestrates while GPA handles the GUI execution. We conducted a pilot experiment to compare GPA with Gemini 3 Pro (with CUA tools) and found that GPA achieves higher success rate with $10\times$ faster execution speed in finishing long-horizon GUI tasks.
    

\end{abstract}

\input{content/1_intro}
\input{content/2_method}
\input{content/3_experiments}
\input{content/4_related_work}
\input{content/5_conclusions}

\bibliographystyle{plain}
\bibliography{reference.bib}
\input{content/6_appendix}
\end{document}

%% file: content/1_intro.tex
\section{Introduction}

The automation of repetitive tasks on Graphical User Interfaces (GUIs) is a critical driver of enterprise productivity. Historically, this domain has been dominated by Robotic Process Automation (RPA). While effective, traditional RPA imposes a heavy implementation burden: it typically requires skilled developers to manually define selectors (e.g., HTML tags, accessibility IDs) and script logic to handle edge cases. Furthermore, these scripts are notoriously brittle; a minor update to a website’s layout or a change in screen resolution can break the rigid coordinate or metadata-based rules, requiring costly maintenance~\citep{haerens2020evolvability,eikebrokk2020rpa}. Beyond fragility, many tasks remain out of reach for RPA entirely: workflows involving dynamic content, complex conditional logic, or unstructured visual data cannot be expressed as deterministic rules.

Recently, large Vision Language Model (VLM)-based GUI agents have emerged as a more flexible alternative~\citep{wang2024guiagentssurvey,tang2025mllmguiagents}. These agents can interpret high-level instructions and interact with diverse interfaces. However, their reliance on generative probability introduces a fundamental flaw for mission-critical workflows: they are non-deterministic. The stochastic nature of probabilistic next-token prediction means that an agent may perform correctly nine times but ``hallucinate'' an action on the tenth. For real-world enterprise use cases that prioritise reliability, this inherent uncertainty is difficult to bound in practice. Additionally, VLM agents offer limited controllability: the same high-level instruction can yield different action sequences across runs, making it hard to audit, constrain, or predict behaviour. They also often suffer from high latency and data privacy risks, as they typically require streaming sensitive screenshots to external cloud-based APIs for inference~\citep{anthropic2024computeruse,openai2025cua,google2025geminicu}.

\begin{table*}[h]
\centering
\small\sffamily
\setlength{\tabcolsep}{8pt}
\renewcommand{\arraystretch}{1.35}
\renewcommand{\tabularxcolumn}[1]{m{#1}}
\setlength{\aboverulesep}{0pt}
\setlength{\belowrulesep}{0pt}
\begin{tabularx}{\textwidth}{>{\bfseries}m{2.2cm} >{\centering\arraybackslash\columncolor{myblue!8}}X >{\centering\arraybackslash}X >{\centering\arraybackslash}X}
\toprule
\rowcolor{gray!10}
\cellcolor{gray!10} & \cellcolor{myblue!18}\textbf{GPA} & \textbf{GUI Agents} & \textbf{Traditional RPA} \\
\specialrule{\lightrulewidth}{0pt}{0pt}
Setup & Learn from a single demonstration & Prompt engineering & Manual scripting and selector labelling \\
\rowcolor{gray!10}
\cellcolor{gray!10} Runtime AI & \cellcolor{myblue!18} Tiny local models only & LLM call per action & No AI \\
Reliability & Deterministic, reliable replay & Non-deterministic & Fragile under UI drift \\
\rowcolor{gray!10}
\cellcolor{gray!10} Privacy & \cellcolor{myblue!18} Fully local & Screenshots sent to external providers & Local \\
Variables & Auto-extracted at build time & Ad hoc via prompting & Hard-coded \\
\bottomrule
\end{tabularx}
\caption{\textbf{Comparison of GPA with VLM-based GUI agents and traditional RPA}. GPA combines demonstration-based workflow construction with deterministic local execution.}
\label{tab:intro_comparison}
\end{table*}

In this work, we present GUI Process Automation (GPA), a system that combines the ease of demonstration-based learning with robust geometric matching. Unlike RPA, GPA requires no coding or selector inspection; the user records a single demonstration of the task, from which the system compiles the interaction trace into a structured workflow with ordered steps, action types, neighbor nodes, and parameterisable fields. Unlike VLM agents, GPA executes this workflow deterministically and efficiently, runs entirely locally to preserve data privacy, and is safeguarded by readiness checking---a confidence-gating mechanism that verifies the UI state before each action is executed.

The main technical challenge is then \emph{UI grounding}~\citep{cheng2024seeclick}: given a recorded workflow step, reliably locating the corresponding interactive element on the current screen despite layout changes. Appearances shift due to rendering updates, dynamic content, and window rescaling, while element identifiers may be absent or inconsistent across sessions. Addressing this robustly requires moving beyond pixel coordinates or fragile metadata lookups. In summary, the key contributions of GPA are:

\begin{itemize}
    \item \textbf{Robustness via Geometric Context}: We employ a stable recording and building process with a multi-stage retrieval process culminating in a context-guided Sequential Monte Carlo (SMC)~\cite{del2006sequential, djuric2003particle} inference procedure. If the primary detector misses the target element, or if the application window is rescaled, the system uses neighboring nodes from the demo graph to infer the target's location. This mimics human intuition, in which a user locates a button based on its position relative to surrounding text or icons.
    \item \textbf{Reliability}: By utilising statistical calibration and a pre-computed null distribution, GPA executes actions only when confidence exceeds a strictly defined threshold. This eliminates the random failures associated with generative AI models and safeguards the system. 
    \item \textbf{Privacy and Performance}: The architecture relies on lightweight local models (e.g., IconCLIP~\cite{iconclip}) rather than massive cloud models. We also trained an accurate UI detector to identify interactive elements. Together, these components ensure high-quality, low-latency execution and guarantee that sensitive visual data never leaves the local machine. In a pilot study, GPA achieves $100\%$ success while running roughly $10\times$ faster on average than Gemini's computer-use agent baseline.
\end{itemize}

%% file: content/2_method.tex
\section{GPA: GUI Process Automation}

A \emph{GUI workflow} or \emph{GUI RPA} is a sequence of GUI actions, such as clicks, text inputs, and hotkeys, performed in a fixed order to accomplish a repeatable task. GPA operates in two phases: (1) a \emph{demonstration phase}, in which a user performs the task once while the system records each (screenshot, action) pair, and (2) an \emph{execution phase}, in which the system automatically replays the workflow with new inputs. We address the UI grounding challenge by representing each UI state as a graph and formulating grounding as a graph-matching problem that combines visual similarity, geometric relationships, and statistical calibration.

\begin{figure}[htbp]
    \centering
    \begin{subfigure}{0.7\textwidth}
        \centering
        \includegraphics[width=\textwidth]{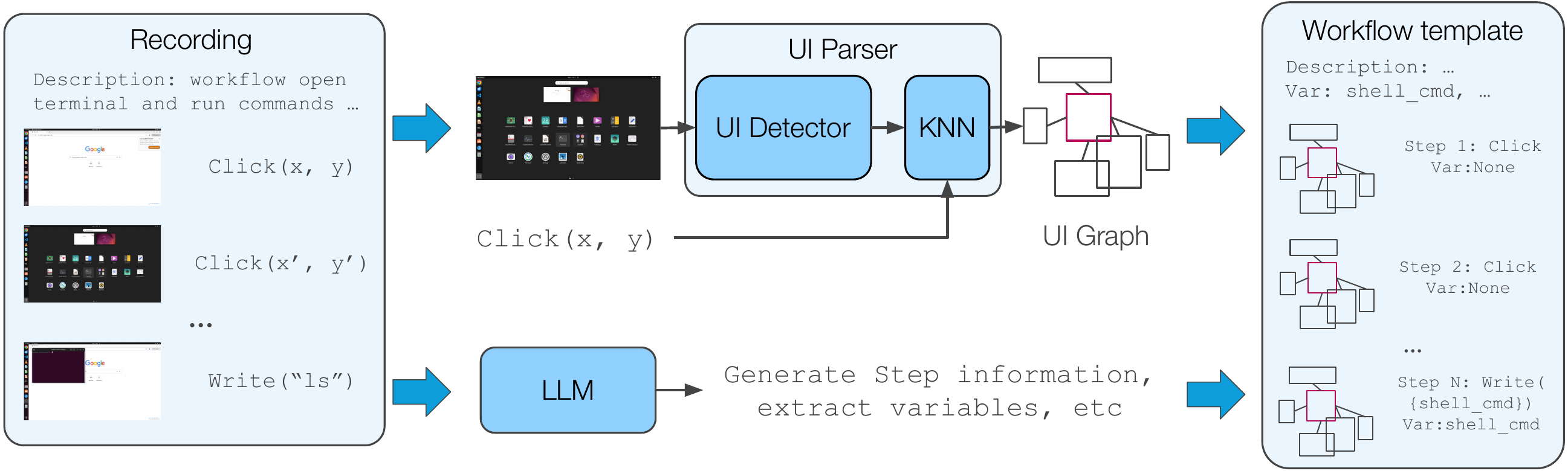}
        \caption{Demonstration phase}
        \label{fig:gpa_build}
    \end{subfigure}
    \vspace{1em}
    \begin{subfigure}{0.27\textwidth}
        \centering
        \includegraphics[width=\textwidth]{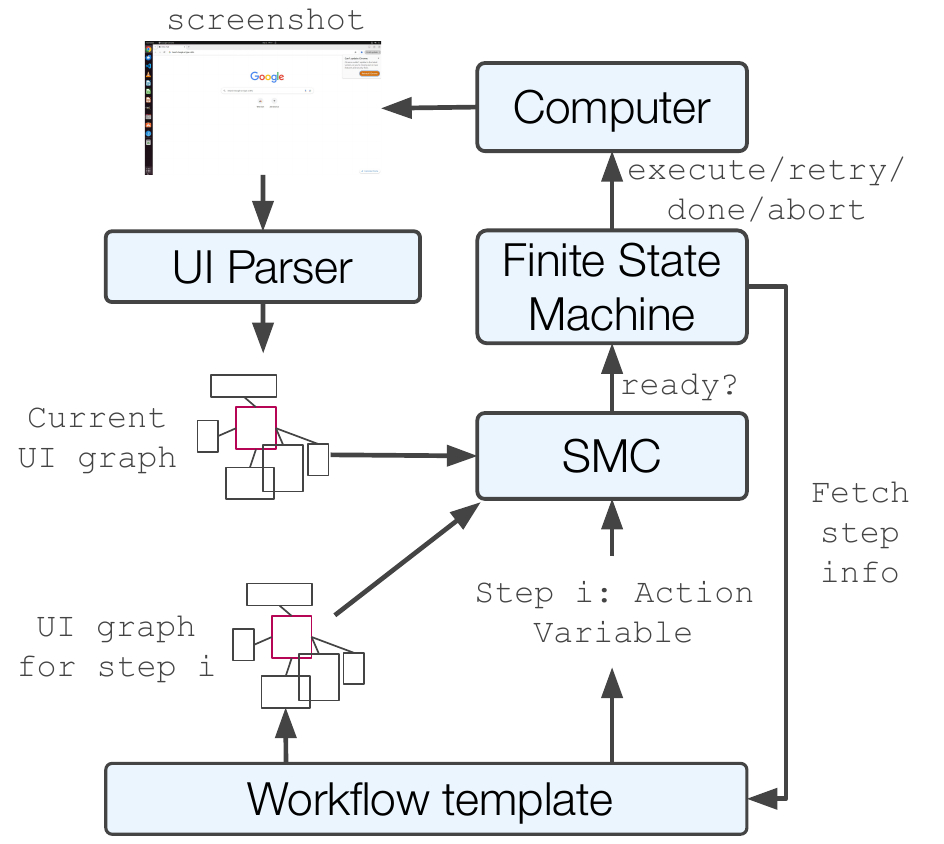}
        \caption{Execution phase.}
        \label{fig:gpa_execute}
    \end{subfigure}
    \caption{\textbf{Two phases of GPA: demonstration phase (a) and execution phase (b)}. In the demonstration phase, a user interacts with the local desktop environment while the recorder captures screenshots and actions, parses them into step subgraphs, and applies LLM-based post-processing for cleanup and variable extraction. In the execution phase, GPA receives the latest observation from the local computer, loads the current recorded step, parses the screenshot into a UI graph, and combines the parsed graph, recorded workflow graph, and action in SMC localization. A finite state machine orchestrates all the components, checking readiness of each step, control execution, and handle errors.}
    \label{fig:gpa_overview}
\end{figure}

As shown in Figure~\ref{fig:gpa_overview}, in the demonstration phase, the system records one user run as a sequence of screenshot-action pairs. For each step, it parses the screenshot into UI elements, builds a UI graph, and stores the target element together with nearby nodes in a reusable workflow template.
In the execution phase, GPA processes the workflow step by step. For each step, it builds a UI graph from the current screen and finds the target element from the demonstration. A finite state machine orchestrates all the components, checking readiness of each step, control execution, and handle errors.

\subsection{Problem Formulation}\label{sec:formulation}

We formulate UI grounding as a localization problem, relying on two assumptions. First, nearby UI elements often preserve their relative layout across runs, even when the window is resized or slightly rearranged. Second, even if the target itself changes appearance or is missed by the detector, some nearby elements usually remain stable enough to serve as context for geometric inference.

We represent each UI state as a graph $G=(V,E)$, where each node $v\in V$ stores a bounding box $b_v$, OCR text $t_v$, and an icon embedding $\mathbf{e}_v$. Edges connect spatially nearby elements. For each demonstrated action, we form a demo graph $G_d$ consisting of the target node $v_\mathrm{target}$ and its nearby nodes $\{v_i\}_{i=1}^M$. Given a runtime graph $G_r$, the goal is to predict the click location of the target on the current screen.

\begin{wrapfigure}{r}{0.34\textwidth}
    \centering
    \vspace{-0.8em}
    \includegraphics[width=0.34\textwidth]{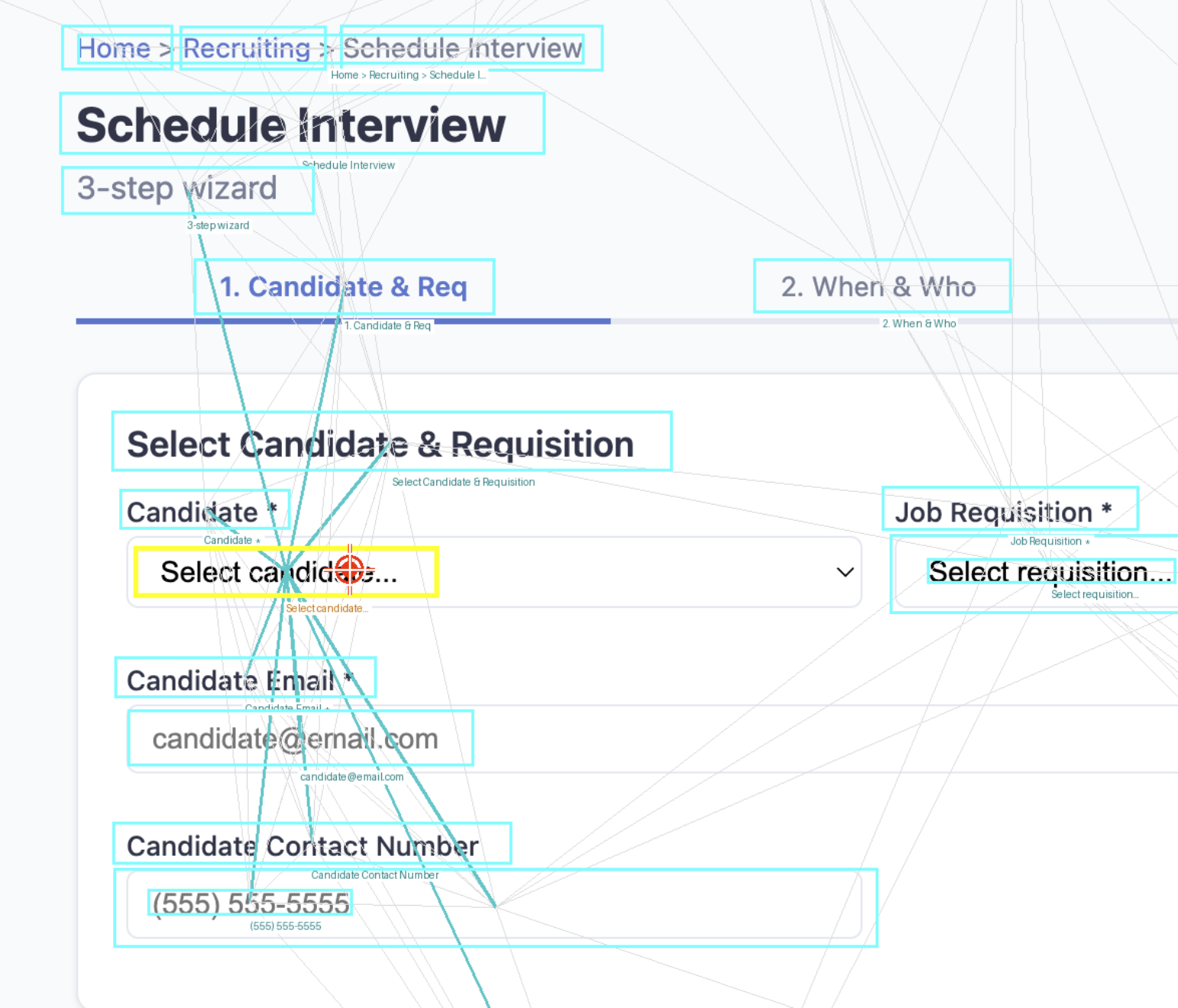}
    \caption{\textbf{Example UI graph with detected elements and edges}. The highlighted node is the target, and nearby nodes helps localizing the target.}
    \label{fig:ui_graph_example}
\end{wrapfigure}
\begin{figure*}[t]
    \centering
    \includegraphics[width=\linewidth]{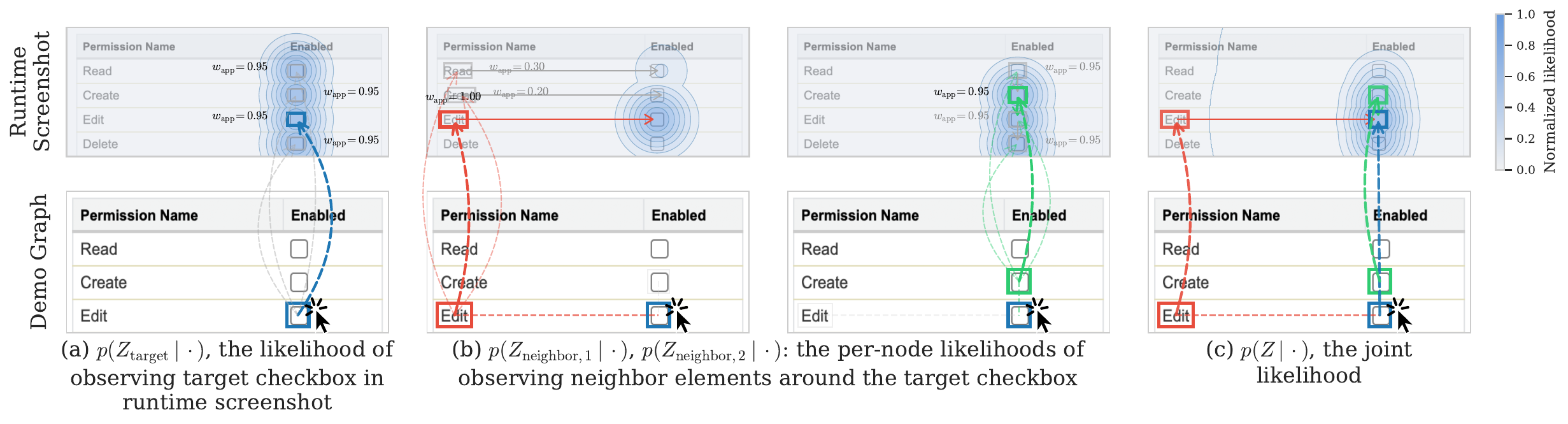}
    \caption{\textbf{SMC likelihood visualization}. The target to click is the checkbox next to ``Edit''. (a) shows the likelihood of directly matching the target checkbox. The pointwise maximum over candidates forms the likelihood surface, showing high ambiguity since matching candidates are all identical. (b) shows the likelihood of matching the neighbor nodes of target, which helps us to differentiate the checkbox. (c) shows the joint likelihood, combining the results of (a) and (b).}
    \label{fig:likelihood_vis} 
\end{figure*}

To handle layout changes, we infer a latent variable $\theta = [x, y, s_x, s_y]$, where $(x,y)$ is the target location and $(s_x, s_y)$ models horizontal and vertical scaling between the demonstration and runtime screens\footnote{We use execution screen, runtime screen or current screen interchangeably.}. 
Let $Z_{\text{target}}$ be the event of observing the target node in the runtime screenshot, and $Z_{\text{neighbor},i}$ be observing the matching instance of the target's $i$-th neighbor in the runtime screenshot at the same geometric relation to the target. $Z = Z_{\text{target}} \cup Z_{\text{neighbor},1} \cup Z_{\text{neighbor},2} \cup ...$ denote the overall observation event across all neighbor nodes. For example, in Figure~\ref{fig:likelihood_vis}, the $Z_{\text{target}}$ denotes observing the matching instance of target checkbox, where $Z_{\text{neighbor},1}$ denotes observing the ``Edit'' text next to the target checkbox at the runtime screenshot.
We seek the posterior
\begin{equation}
p(\theta \mid Z) \propto p(Z \mid \theta)\, p(\theta),
\end{equation}
where $p(\theta)$ is a prior over plausible locations and scales, and $p(Z \mid \theta)$ is the likelihood of the observing the same demo graph in the runtime screenshot under hypothesis $\theta$.
Exact inference is difficult because context-candidate correspondences are ambiguous and the posterior can be multi-modal. We therefore use Sequential Monte Carlo (Section~\ref{sec:smc})~\cite{del2006sequential, djuric2003particle} to approximate the posterior and obtain a click prediction with a confidence score for readiness check.

The UI graph is constructed from screenshots in three steps: (1)~extracting UI elements using a finetuned icon detector\footnote{The model is available at \url{https://huggingface.co/Salesforce/GPA-GUI-Detector}} and OCR for text, (2)~computing visual features for each element using IconCLIP~\cite{iconclip}, and (3)~connecting spatially nearby elements via $k$-nearest neighbors to form edges. The resulting graph captures both visual appearance and spatial layout, allowing the retrieval algorithm to exploit neighboring context when matching across UI states. See Appendix~\ref{sec:appendix_ui_graph} for details.

We compute similarity between nodes based on their type. For textual elements, we combine fuzzy string matching with icon embedding similarity; for non-textual elements (icons, images), we use icon embedding similarity alone. This hybrid approach is robust to both visual changes and OCR errors (see Appendix~\ref{sec:appendix_similarity} for the full formulation).

\subsection{Demonstration and Workflow Building}

During the demonstration phase, the user performs the task once while the recorder captures each (screenshot, action) pair as a keyframe. A background preprocessing thread immediately parses each incoming screenshot into a UI graph as keyframes arrive, so that processing overlaps with the user's interaction and the workflow is ready to compile as soon as the recording ends.

For each step, GPA identifies the target UI element at the recorded click coordinates and selects nearby nodes using $k$-nearest-neighbour graph traversal, forming a \emph{step subgraph} that encodes the target together with its local neighbors. For steps following a scroll action, the screenshot is preprocessed to mask regions outside the scrollable container before node matching, ensuring that neighbor nodes are drawn from within the scrolled area rather than from unrelated parts of the screen.

After recording, an LLM analyses the full step sequence to assign natural-language descriptions, identify parameterisable fields (e.g., form values, search terms) as typed workflow variables, and generate a workflow name and title. The result is a \emph{workflow template}---a structured file containing the ordered steps, each with its step subgraph, action type, and variable bindings. At execution time, the runner loads this template and replays it against the live UI (see Appendix~\ref{sec:appendix_workflow_example} for an example).

\subsection{Execution}

At runtime, GPA processes the workflow one step at a time. For each step, the runner captures an observation (screenshot and window metadata) from the environment, parses it into a UI graph, retrieves the target element from the demonstration subgraph, and checks whether the current screen is ready to act. If the confidence score exceeds a threshold, the action is generated and executed; otherwise the step is retried. To reduce per-step latency, a lightweight precheck pipeline speculatively processes the next step in the background while the environment is settling after the current action.

\paragraph{Grounding as Sequential Monte Carlo Localization}\label{sec:smc}

We first attempt a direct match using appearance similarity between the target node $v_t$ in the demo graph and nodes in runtime graph $G_r$. We rank matching candidates by visual and textual similarity and return the top candidate when it is both \emph{high-confidence} and \emph{unambiguous}. Ambiguity is measured by the normalised entropy of the softmax distribution over the top-$k$ candidate scores (see Appendix~\ref{sec:appendix_entropy}): low entropy indicates a clear winner, while high entropy triggers the full SMC procedure. This fast path handles the common case where the UI is largely unchanged and avoids running SMC unnecessarily.

Hard cases arise when target matches are ambiguous or having low confidence, motivating us to use neighboring nodes in the demo graph to help locate the target element. We solve for $\theta$ via the posterior defined in Section~\ref{sec:formulation}. Given a demo graph, we can get target coordinate $\mathbf{t}_{\text{demo}}$ and neighbor node coordinates $\{\mathbf{c}_i^{\text{demo}}\}_{i=1}^{M}$, and precompute neighbor-to-target displacement vectors $\mathbf{r}_i = \mathbf{c}_i^{\text{demo}} - \mathbf{t}_{\text{demo}}=[r_{ix},r_{iy}]^\top$.
For each point $\theta$, the predicted neighbor node location on the current screen is $\hat{\mathbf{c}}_i(\theta) = [x, y] + [s_x r_{ix},\, s_y r_{iy}]$.

For each node $v\in V_\mathrm{demo}$ in demo graph, we can find a set of matching candidates in the runtime graph $\mathcal{C}_v$ obtained by appearance filtering (text match when available, otherwise icon embedding similarity). Instead of committing to a single correspondence, we jointly consider all candidates with explicit missing-node mass. For each neighbor node, we take the \emph{best-matching} candidate (weighted Gaussian) and combine it with the missing mass:
\begin{wrapfigure}{r}{0.27\textwidth}
    \centering
    \begin{tikzpicture}
    \begin{axis}[
        xmin=0.1, xmax=3.2,
        ymin=0, ymax=1.0,
        grid=none,
        xlabel={$s$},
        ylabel={$p(s)$},
        every tick label/.append style={font=\scriptsize},
        every axis label/.append style={font=\footnotesize},
        every axis x label/.style={at={(ticklabel* cs:1.0)}, anchor=west},
        every axis y label/.style={at={(ticklabel* cs:1.0)}, anchor=south},
        xtick={1,2.2},
        xticklabels={$1$, $\exp(\mu)$},
        ytick=\empty,
        scale=1.0,
        width=\linewidth,
        height=0.8\linewidth
    ]
        \addplot[myblue, very thick, samples=500, smooth, domain=0.001:3.2]
        {0.4*(1/(x*0.4*sqrt(2*pi)))*exp(-((ln(x)-0)^2)/(2*0.4^2))
        + 0.6*(1/(x*0.2*sqrt(2*pi)))*exp(-((ln(x)-ln(2.2))^2)/(2*0.2^2))};
        \addplot[myred, dashed, thick] coordinates {(1,0) (1,1.45)};
        \addplot[myred, dashed, thick] coordinates {(2.2,0) (2.2,1.45)};
    \end{axis}
    \end{tikzpicture}
    \caption{\textbf{Mixture prior over scale $s$}: one mode favors no resizing ($s=1$), and the other favors proportional rescaling.}
    \label{fig:lognormal_mixture}
\end{wrapfigure}
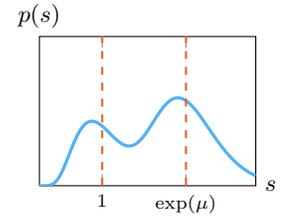
\begin{equation}
p(Z_v \mid \theta) = p_{\text{miss}} + \max_{c \in \mathcal{C}_v} \left( w_{\text{app}}(v,c)\;
\mathcal{N}\!\left(\mathbf{p}_c \mid \hat{\mathbf{c}}_v(\theta), \sigma_v^2 I\right) \right),
\end{equation}
where $\mathbf{p}_c$ is the matching candidate center, $w_{\text{app}}(n,c)$ is the appearance similarity, and $p_{\text{miss}}$ allows nodes to be absent or unmatched. $\mathcal{N}\!\left(\mathbf{p}_c \mid \hat{\mathbf{c}}_v(\theta), \sigma_v^2 I\right)$ evaluates how likely the matching candidate of node $i$ has the same displacement $\mathbf{r}_n$ to $\theta$ as in the demo graph. The joint likelihood weights neighbor nodes by locality $w_{\text{loc}}(n)$ (closer neighbor nodes are more reliable to infer the location of target UI element; See Appendix~\ref{sec:appendix_locality}) and add a scale prior:
\begin{equation}
\log p(Z\mid\theta) + \log p(\theta)
=
\sum_{v\in V_\mathrm{demo}} w_{\text{loc}}(v)\, \log p(Z_v \mid \theta) + \log p(s_x) + \log p(s_y),
\end{equation}
where the scale prior for each axis is a \emph{mixture of two log-normals} (as shown in Figure~\ref{fig:lognormal_mixture}): one component centered at $s=1$ (no resizing) and another at the observed window-size ratio (proportional rescaling). This allows the model to handle both unchanged and rescaled windows. Full details of the scale prior parameterization and geometric tolerance are in Appendix~\ref{sec:appendix_scale_prior}.

We approximate the posterior $p(\theta\mid Z)$ using a tempered Sequential Monte Carlo (SMC) sampler with $N$ weighted particles. Particles are initialized by back-projecting top-$K$ context-candidate matches into target proposals with Gaussian jitter. The sampler then applies likelihood tempering: gradually ``fading in'' the likelihood from the prior to the full posterior, with adaptive temperature steps, importance reweighting, resampling, and Metropolis--Hastings rejuvenation at each stage. This standard SMC sampler procedure prevents particle degeneracy and handles the multi-modal posterior that arises from ambiguous context-candidate correspondences. The final prediction $\hat{\mathbf{t}}$ is the mean of the densest particle cluster. The likelihood computation, SMC iterations, etc, are all very fast in runtime with vectorized operations (in less than 0.2 sec). Full algorithmic details are in Appendix~\ref{sec:appendix_smc_algorithm}.


\begin{figure}
    \centering
    \includegraphics[width=\linewidth]{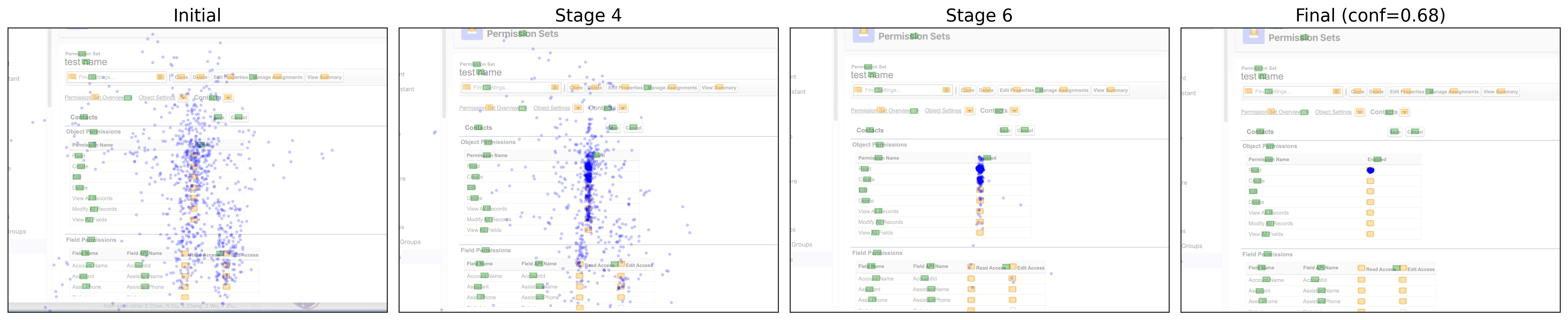}
    \caption{\textbf{Process of SMC}. The blue points are particles that gradually converge to the target prediction. }
    \label{fig:graph_demo}
\end{figure}

\paragraph{Readiness Checking}
Before executing each action, GPA gates on a confidence score $C = \tilde{p}(Z\mid \bm{\theta})\times C_{\text{spatial}}$ derived from the SMC retriever. \emph{Likelihood confidence} $\tilde{p}(Z\mid \bm{\theta})$ measures how well the predicted position explains the observed demo-runtime matches. It is different from the likelihood of individual particles but a overall confidence of clusters. \emph{Spatial confidence} $C_{\text{spatial}}$ measures how tightly particles agree on the target location, based on the posterior predictive variance. The multiplicative combination ensures that both the evidence quality and posterior certainty must be high: strong likelihood with a scattered posterior (ambiguous) or a tight posterior with poor likelihood (confidently wrong) both yield low confidence. GPA proceeds only when $C$ exceeds a threshold; otherwise the step is retried (see Appendix~\ref{sec:appendix_confidence} for details).

\begin{figure}[htbp]
    \centering
    \begin{subfigure}{0.5\textwidth}
        \centering
        \includegraphics[width=\textwidth]{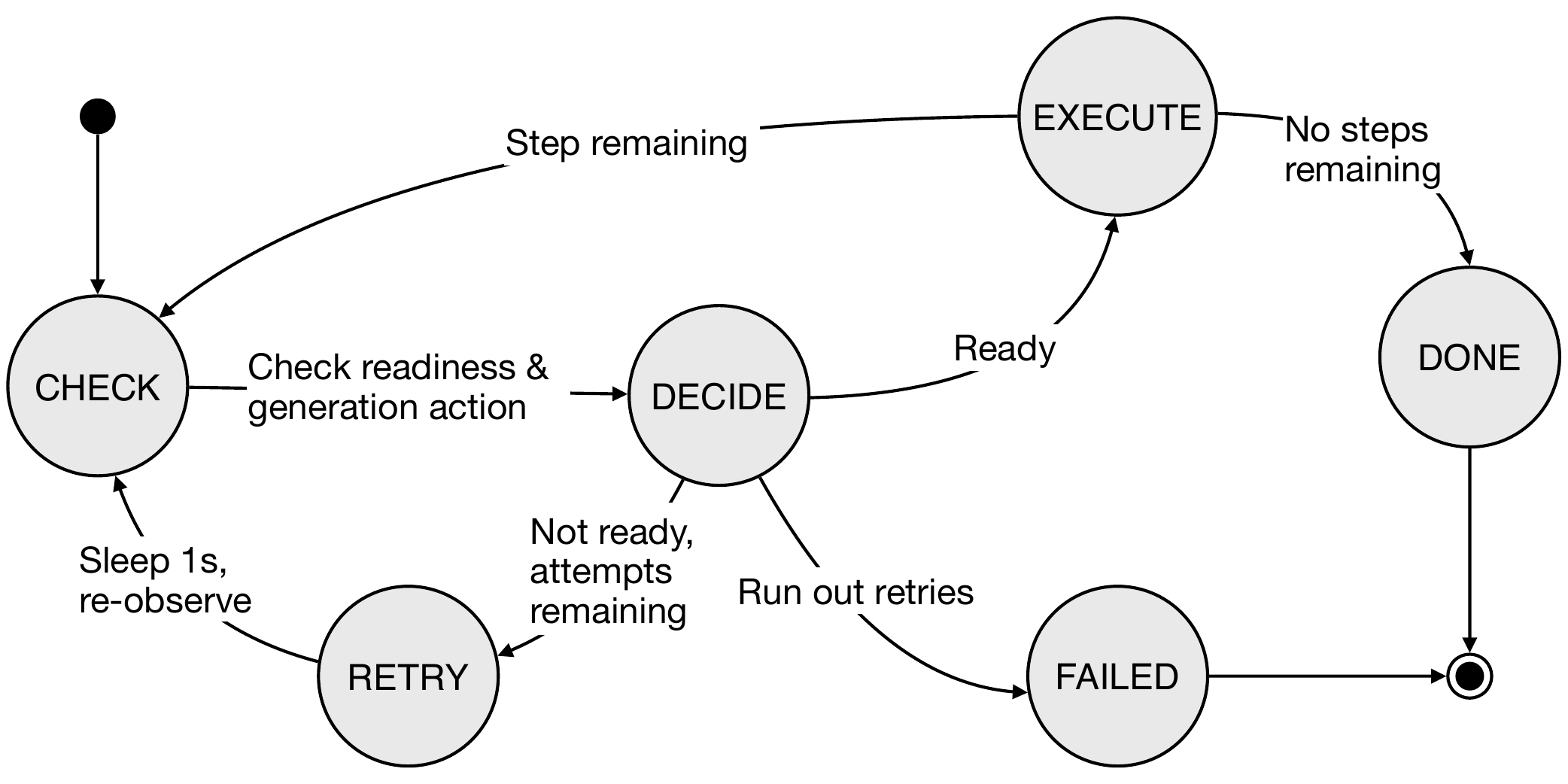}
        \caption{Top-level step-execution FSM.}
        \label{fig:step_fsm_main}
    \end{subfigure}
    \vspace{0.5em}
    \begin{subfigure}{0.49\textwidth}
        \centering
        \includegraphics[width=\textwidth]{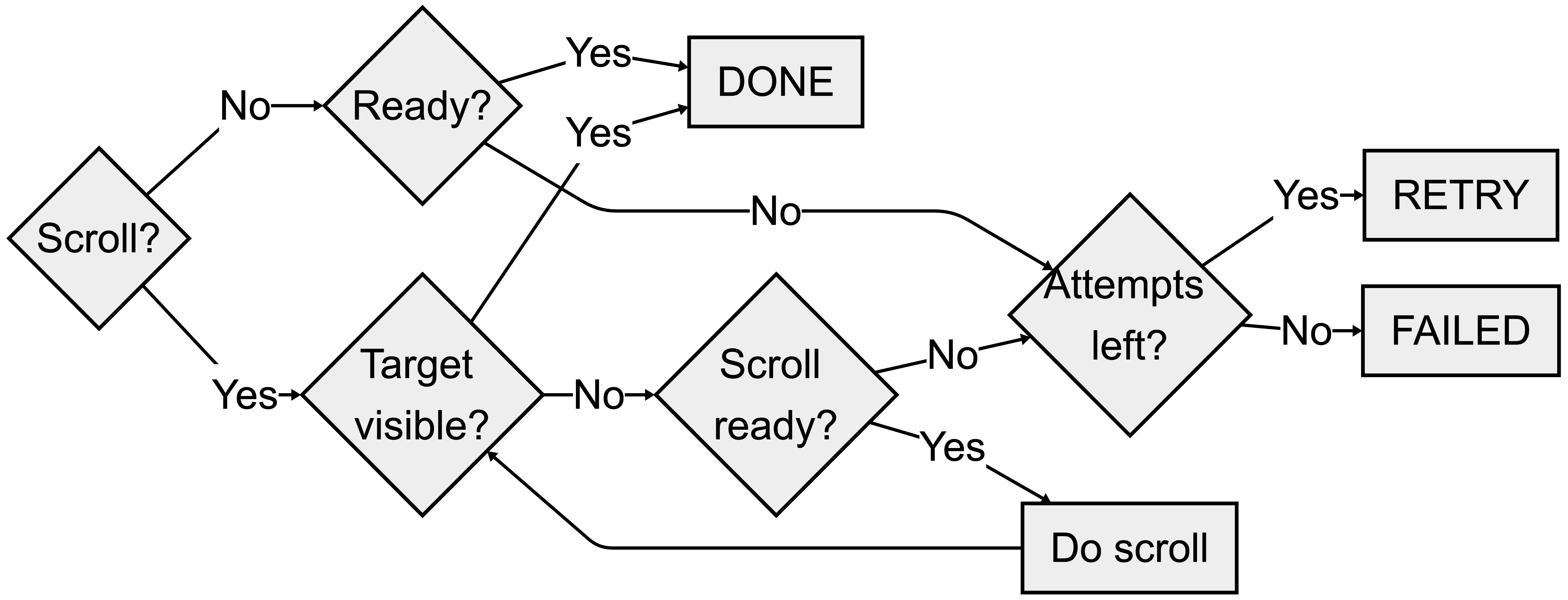}
        \caption{\textsc{Decide} state logic. Scroll-to-find steps first check target visibility; regular steps follow the bounded retry path directly.}
        \label{fig:step_fsm_decide}
    \end{subfigure}
    \caption{\textbf{Step-execution FSM and the \textsc{Decide} state detail}. The top-level FSM is shared across all steps, while the decision logic adds a small scroll-specific branch when needed.}
    \label{fig:step_fsm}
\end{figure}

\paragraph{Execution Control}

Grounding and readiness checking determine \emph{where} GPA should act and \emph{whether} the current screen matches the expected state. A step-level execution controller then decides \emph{what to do next}. As shown in Figure~\ref{fig:step_fsm_main}, the top-level controller follows a simple bounded-retry finite state machine (FSM). For each workflow step, GPA alternates between \textsc{Execute} and \textsc{Decide}. The \textsc{Execute} state parses the current UI, checks readiness, and generates an action candidate. The \textsc{Decide} state then chooses among three outcomes: finish the step, retry after re-observing the screen, or fail the step after exhausting the retry budget. This structure prevents the workflow from drifting after a transient mismatch.

The \textsc{Decide} state contains the only step-specific branch. For regular steps, shown in Figure~\ref{fig:step_fsm_decide}, the logic is minimal: if the action is ready, GPA finishes the step; otherwise, it either retries or fails depending on the remaining attempt budget. A special case is \emph{scroll-to-find step}, which is a scroll action whose purpose is to reveal a later target element that is currently off-screen. For scroll-to-find steps, \textsc{Decide} first checks whether the target is already visible. If so, GPA skips the scroll and completes the step immediately. If the target is not yet visible, GPA checks whether the scroll action itself is ready. A ready scroll action is executed and the next screen is re-evaluated on the following iteration; otherwise, GPA follows the same bounded retry-or-fail rule as in the regular case. In this way, scroll-specific logic refines the decision state without changing the top-level FSM. 

This finite-state controller complements the SMC-based matcher: the matcher provides localization and confidence, while the controller turns that confidence into safe retry, execute, or fail decisions. GPA further reduces latency with a lightweight precheck pipeline that speculatively processes upcoming steps in the background while the environment settles after the current action (see Appendix~\ref{sec:appendix_precheck}).

%% file: content/3_experiments.tex
\section{Pilot Experiments}
\label{sec:pilot_experiments}

We conduct a small-scale pilot study of $16$ desktop GUI tasks to compare GPA against a VLM-based baseline: the Gemini computer-use agent ({\tt gemini-3-pro}) that executes a demonstrated workflow description. We categorize tasks by the \emph{length of the recorded demonstration} (number of steps): \textbf{simple} tasks have short demos and \textbf{hard} tasks have long demos. For each task, we measure (i) wall-clock runtime and (ii) success rate.

The task set covers a mix of common desktop productivity and enterprise workflows. The \textbf{simple} tasks (average $10.8$ steps) include drafting an email, downloading a receipt from email, downloading a utility receipt followed by counting, flight searching, and flight booking. The \textbf{hard} tasks (average $27.27$ steps) include setting a Google Calendar event, setting a calendar event followed by drafting an email, flight searching and booking, two Agentforce tasks, reimbursement submission with and without receipt retrieval, two SAP ERP form-filling tasks, and two HR workflows for interview scheduling and candidate information entry. This mix lets us evaluate both short routine automations and longer multi-stage enterprise processes.
\begin{table}[h]
\centering
\small\sffamily
\setlength{\tabcolsep}{8pt}
\renewcommand{\arraystretch}{1.35}
\renewcommand{\tabularxcolumn}[1]{m{#1}}
\setlength{\aboverulesep}{0pt}
\setlength{\belowrulesep}{0pt}
\begin{tabularx}{\columnwidth}{>{\bfseries}m{2.0cm}!{\vrule width 0.3pt}>{\centering\arraybackslash}X >{\centering\arraybackslash}X >{\centering\arraybackslash}X >{\centering\arraybackslash}X >{\centering\arraybackslash}X}
\toprule
\rowcolor{gray!10}
\cellcolor{gray!10} & & \multicolumn{2}{c}{\textbf{GPA}} & \multicolumn{2}{c}{\textbf{Gemini 3 pro}} \\
\specialrule{\lightrulewidth}{0pt}{0pt}
\rowcolor{gray!10}
\cellcolor{gray!10} Task level & Avg demo steps & Success (\%) & Time (s) & Success (\%) & Time (s) \\
\specialrule{\lightrulewidth}{0pt}{0pt}
\cellcolor{gray!10} Simple & 10.80 & 100 & 17.84 & 93.2 & 210.66 \\
\rowcolor{gray!10}
\cellcolor{gray!10} Hard & 27.27 & 100 & 40.96 & 87.64 & 383.24 \\
\cellcolor{gray!10} Avg & 22.13 & 100 & 33.74 & 89.38 & 329.31 \\
\bottomrule
\end{tabularx}
\caption{\textbf{Pilot results: GPA vs.\ Gemini GUI agent.} Success rate and runtime aggregated by task difficulty (defined by demonstration length).}
\label{tab:pilot_experiments}
\end{table}

\paragraph{Experimental protocol}
For each task, GPA is given a single recorded demonstration and then replays the workflow using its visual graph retrieval, context-node-based matching, readiness checks, and bounded retry logic. The Gemini baseline is given the same demonstrated video and process theml into textual form, and use it to guide its action autoregressively during execution. We report average runtime and success rate over the tasks in each difficulty group. Although this is a pilot-scale study, it is sufficient to test whether the practical trends are consistent with the design goals of GPA. 

\paragraph{Results}
As shown in Table~\ref{tab:pilot_experiments}, GPA achieves $100\%$ success across both difficulty levels while being roughly $10\times$ faster ($33.74$\,s vs.\ $329.31$\,s). Gemini's success drops from $93.2\%$ to $87.64\%$ as tasks grow longer. Three factors explain the gap:

\begin{itemize}[leftmargin=*,itemsep=3pt]
    \item \textbf{Reliability.}
    GPA follows a fixed demonstrated procedure and acts only when the readiness checker confirms a confident match. Bounded retries handle transient delays; unresolvable mismatches trigger explicit failure rather than guessing. Gemini must infer the correct action at every step, and each inference carries a small misidentification risk that compounds over the trajectory.

    \item \textbf{Latency.}
    Each GPA step involves only on-device screenshot capture, OCR parsing, and local subgraph retrieval (milliseconds to low seconds). Gemini incurs network round-trip and VLM inference latency per step, which accumulates to $>$$383$\,s on ${\sim}27$-step hard tasks versus ${\sim}41$\,s for GPA.

    \item \textbf{Scaling with length.}
    GPA's fixed-plan execution adds only a cheap matching operation per step with no additional drift risk. The generative agent may misread a label, click an adjacent element, or miss a page load at each step---over $27$ steps these per-step failure probabilities compound, explaining the ${\sim}6$\,pp success drop while GPA remains at $100\%$.
\end{itemize}

%% file: content/4_related_work.tex
\section{Related Work}

\textbf{From RPA to visual GUI automation.}
Classical robotic process automation (RPA) is designed for stable, repetitive workflows and is attractive in practice because it can automate legacy systems without requiring API-level integration. At the same time, prior work has emphasized that RPA is most effective when processes are structured, rule-based, and relatively invariant, and that coupling automation logic to application structure or surface behavior makes such systems brittle under software evolution and interface changes \citep{haerens2020evolvability, eikebrokk2020rpa}. Earlier work in visual GUI automation and testing partially addressed this problem by replacing DOM- or selector-centric interaction with screenshot matching \citep{chang2010gui} and image-based scripting \citep{alegroth2015visual}. However, these methods still rely on low-level visual correspondence and offer limited semantic understanding, which leaves them sensitive to appearance changes and ambiguous interface states.

\textbf{Foundation-model GUI agents.}
Recent work reframes GUI automation as a multimodal decision-making problem over screenshots, structured interface cues, and grounded actions. Recent surveys organize the space by perception, grounding, planning, memory, and execution across web, mobile, and desktop settings \citep{wang2024guiagentssurvey, tang2025mllmguiagents}. The line first matured on the web through datasets and evaluation suites such as Mind2Web \citep{deng2023mind2web}, WebArena \citep{zhou2023webarena}, and VisualWebArena \citep{koh2024visualwebarena}, which exposed the gap between text-only agents and visually grounded multimodal agents. End-to-end web agents such as WebVoyager \citep{he2024webvoyager} and AutoWebGLM \citep{lai2024autowebglm} showed that large multimodal models can act directly on live websites, while SeeClick and ScreenSpot \citep{cheng2024seeclick} highlighted GUI grounding as a central bottleneck for more general visual interaction. Subsequent environments expanded evaluation from browser automation to full computer use, including OmniACT \citep{kapoor2024omniact}, OSWorld \citep{xie2024osworld}, AndroidWorld \citep{rawles2024androidworld}, WindowsAgentArena \citep{bonatti2024windowsagentarena}, and ScreenSpot-Pro \citep{li2025screenspotpro}, shifting attention toward long-horizon, cross-application tasks under realistic interface variability.

At the system level, early foundation-model GUI agents such as CogAgent \citep{hong2023cogagent}, Ferret-UI \citep{you2024ferretui}, AppAgent \citep{zhang2023appagent}, Mobile-Agent \citep{wang2024mobileagent}, and UFO \citep{zhang2024ufo} established screenshot-native interaction for web, mobile, and Windows environments. More recent work has diversified into stronger open-source grounding/action models, including UGround \citep{gou2024uground}, OS-ATLAS \citep{wu2024osatlas}, Aguvis \citep{xu2024aguvis}, and TinyClick \citep{pawlowski2024tinyclick}, and into increasingly capable execution frameworks such as AutoGLM \citep{liu2024autoglm}, Agent~S \citep{agashe2024agents}, Agent~S2 \citep{agashe2025agents2}, UI-TARS \citep{qin2025uitars}, PC-Agent \citep{liu2025pcagent}, Mobile-Agent-v3 \citep{ye2025mobileagentv3}, Mobile-Agent-v3.5 \citep{xu2026mobileagentv35}, AppAgentX \citep{jiang2025appagentx}, and UFO2 \citep{zhang2025ufo2}, which combine hierarchical planning, specialist modules, reinforcement learning, memory, or hybrid GUI/API control to improve long-horizon execution. In parallel, productized computer-use systems such as Anthropic's computer use \citep{anthropic2024computeruse}, OpenAI's Operator \citep{openai2025operator} and CUA \citep{openai2025cua}, Google's Project Mariner \citep{google2024projectmariner} and Gemini 2.5 Computer Use \citep{google2025geminicu}, and open-source platforms such as OpenClaw \citep{openclaw2026} show that GUI agents are rapidly moving from benchmark prototypes toward deployable assistants and automation substrates.

\textbf{Learning from demonstration.}
Another major direction reduces the burden on zero-shot planning by using demonstrations to specify the desired workflow. Early work such as HILC \citep{intharah2017hilc} showed that users can teach GUI tasks through demonstration, with follow-up questions helping resolve ambiguity in one-shot task specification. More recently, LearnAct \citep{liu2025learnact} introduces a demonstration-based framework for mobile GUI agents together with the LearnGUI benchmark, using dedicated modules to parse demonstrations, retrieve relevant prior experience, and execute actions in new contexts. Instruction Agent \citep{li2025instructionagent} similarly leverages a single expert demonstration to extract step-by-step instructions and constrains execution with verification and backtracking, improving performance on complex long-horizon tasks. This literature is especially relevant for settings where the goal is not unconstrained autonomy, but dependable reproduction of user-preferred workflows.



%% file: content/5_conclusions.tex
\section{Conclusions \& Limitations}

We propose GPA, a novel RPA framework that requires only a single demonstration. Powered by mature algorithms for grounding UI elements during process replay, we show that many GUI operations may not require expensive LLM agents to perform. GPA can serve as a powerful tool for GUI automation on local desktops, in browsers, in enterprise software, and in other domains where GUI elements are well-defined and stable. It can also be viewed as a computer-use skill with command-line (CLI) support, enabling integration with shell scripting and composition of multiple workflows for even more complex and flexible use cases.

We also acknowledge several limitations. GPA is a record-and-replay system: it does not have reasoning or decision-making capabilities. It executes workflows as recorded and cannot adapt to situations that require judgment. For example, selecting a date on a calendar widget requires reasoning about the current state (e.g., which month is currently displayed, how many clicks are needed to navigate to the target month). GPA cannot perform this reasoning, so date pickers will only work correctly if the same date as in the recording is being selected.

GPA can be extended to support fully automated, no human-in-the-loop operation: LLM agents could record workflows and perform self-healing when a workflow becomes stale due to UI updates. We are also interested in precondition tracking for tool use, as the preconditions required for a tool call may not always be satisfied, requiring we have a reasonable state estimation of the software we are operationg on. This represents a general challenge for tool design in AI agent systems.

%% file: content/6_appendix.tex
\newpage
\appendix

\section{SMC Retriever Details}

\subsection{Locality Weight Bandwidth}\label{sec:appendix_locality}

Each context node~$i$ contributes to the log-likelihood weighted by
\begin{equation}
w_{\text{loc}}(i) = \exp\!\left(-\frac{\|\mathbf{v}_i\|^2}{2\,\sigma_{\text{loc}}^2}\right),
\end{equation}
where $\mathbf{v}_i = \mathbf{c}_i^{\text{demo}} - \mathbf{t}_{\text{demo}}$ is the demo displacement vector. The bandwidth $\sigma_{\text{loc}}$ controls how rapidly the weight decays with distance from the target: a small $\sigma_{\text{loc}}$ trusts only nearby nodes, while a large one incorporates distant ones.

\paragraph{Motivation.}
In dense interfaces (e.g.\ a spreadsheet), nearby nodes are abundant and distant ones are unreliable because the layout may differ across regions. In sparse interfaces (e.g.\ a dialog with a few buttons), even distant nodes provide useful geometric constraints. An effective $\sigma_{\text{loc}}$ must therefore adapt to the spatial distribution of nodes.

\paragraph{Adaptive bandwidth.}
We apply Silverman's rule~\cite{silverman1986density} using the \emph{RMS distance from the target} to context node centres as the scale estimate:
\begin{equation}
\sigma_{\text{loc}} = \text{clamp}\!\left(1.06\;\hat\sigma\;n^{-1/5},\;\sigma_{\min},\;\sigma_{\max}\right),
\end{equation}
where $\hat\sigma = \sqrt{\frac{1}{n}\sum_{i=1}^{n}\|\mathbf{v}_i\|^2}$ is the root-mean-square (RMS) distance from the target to the $n$ demo nodes---i.e., the square root of the average squared displacement magnitude. Unlike the per-axis standard deviation (which is translation-invariant and only measures the spread of context node positions regardless of the target), RMS distance is \emph{target-centric}: it reflects how far nodes actually are from the target. When the target sits at the edge of the context node cluster, $\hat\sigma$ is larger and distant nodes receive more weight; when the target is surrounded by nodes, $\hat\sigma$ is smaller and the locality weight is more selective. The $n^{-1/5}$ factor comes from Silverman's asymptotically optimal bandwidth rate; it means that with more nodes we can afford to be more local.

\paragraph{Behaviour.}
\begin{itemize}[nosep]
\item \emph{Dense UI} (20 nodes within $\sim$100\,px of target): RMS distance $\hat\sigma \approx 70$\,px, $\sigma_{\text{loc}} \approx 1.06 \times 70 \times 20^{-0.2} \approx 41$\,px. Only nearby nodes contribute significantly.
\item \emph{Sparse UI} (3 nodes, avg.\ 500\,px from target): $\hat\sigma \approx 500$\,px, $\sigma_{\text{loc}} \approx 1.06 \times 500 \times 3^{-0.2} \approx 425$\,px. Even distant nodes receive meaningful weight.
\item \emph{Target at edge} (same nodes as dense, but target far from cluster centre): RMS distance increases, giving a larger $\sigma_{\text{loc}}$ so nodes in the far part of the cluster are not prematurely down-weighted.
\item \emph{Edge cases}: fewer than 2 nodes $\Rightarrow$ fallback constant ($\sigma_{\text{loc}} = 1000$\,px); all nodes at target position $\Rightarrow$ $\hat\sigma \to 0$, clamped to $\sigma_{\min} = 30$\,px.
\end{itemize}

\subsection{Entropy-Based Ambiguity Detection}\label{sec:appendix_entropy}

The fast-path direct match bypasses the SMC particle filter when the top candidate is clearly the best. This decision requires an ambiguity measure over the ranked candidate scores.

\paragraph{Normalised entropy.}
Given the top-$k$ candidate similarity scores $s_1 \geq s_2 \geq \cdots \geq s_k$, we form a softmax distribution at temperature~$\tau$ and compute the normalised entropy:
\begin{equation}
H = -\frac{1}{\log k_{\text{eff}}}\sum_{j=1}^{k} p_j \log p_j, \qquad p_j = \frac{\exp(s_j / \tau)}{\sum_{\ell=1}^{k}\exp(s_\ell / \tau)}.
\end{equation}
The normalisation uses the \emph{effective} candidate count $k_{\text{eff}} = \max(|\{j : p_j > 0.01\}|,\, 2)$ rather than the raw~$k$. This prevents dilution from irrelevant tail candidates: when only the top two candidates carry meaningful probability mass, $k_{\text{eff}}=2$ regardless of how many weak candidates exist. Without this guardrail, normalising by $\log k$ (e.g.\ $k=5$) would reduce $H$ and allow ambiguous top-2 pairs to pass as ``clear'' simply because three negligible candidates inflate the denominator. A match is deemed unambiguous when $H < H_{\text{thr}}$.

\paragraph{Role of temperature.}
The temperature $\tau$ controls sensitivity to score differences. Since similarity scores lie in $[0, 1]$, a small $\tau$ (e.g.\ $0.02$) amplifies even modest gaps into sharply peaked distributions, while a large $\tau$ (e.g.\ $0.5$) flattens them. At $\tau = 0.02$:
\begin{itemize}[nosep]
\item Scores $[0.95, 0.40, 0.30]$: $H \approx 0.00$ (unambiguous).
\item Scores $[0.92, 0.90, 0.30]$: $H \approx 0.84$ (ambiguous --- the top two are too close).
\item Scores $[0.85, 0.84, 0.83, 0.82]$: $H \approx 0.90$ (ambiguous --- four near-equal candidates).
\end{itemize}

\paragraph{Comparison with score-gap heuristic.}
The simpler margin test $s_1 - s_2 > \Delta$ only examines the top two candidates and is sensitive to score scale (a gap of $0.06$ means different things at $s_1 = 0.95$ vs.\ $s_1 = 0.50$). Entropy naturally accounts for (i)~the number of close competitors, not just the runner-up; (ii)~the relative magnitudes of all top-$k$ scores; and (iii)~the overall peakedness of the distribution. For instance, a candidate leading by $0.05$ over three close competitors ($H$ high) is more ambiguous than one leading by $0.05$ over a single runner-up with all others far behind ($H$ low), a distinction the margin test cannot make.

\paragraph{Two-stage gating.}
The direct match requires both a minimum score threshold ($s_1 > s_{\min}$, default $0.9$) \emph{and} low entropy ($H < H_{\text{thr}}$, default $0.5$). The score threshold ensures the top candidate is genuinely a good match; the entropy threshold ensures it is clearly better than alternatives. Only when both conditions hold does GPA skip SMC.

\subsection{Confidence Scoring}\label{sec:appendix_confidence}

The final confidence $C = \tilde{p}(Z\mid \bm{\theta}) \times C_{\text{spatial}}$ is the product of two independent measures that capture complementary aspects of retrieval quality.

\subsubsection{Likelihood Confidence}\label{sec:appendix_ctx_conf}

Likelihood confidence measures how well the predicted target position explains the observed node-candidate matches under the likelihood model.

\paragraph{Per-node score.}
For each node~$v\in V_\mathrm{demo}$, define three log-scale quantities evaluated at the predicted hypothesis $\hat\theta$:
\begin{itemize}[nosep]
\item $\log p_{\text{match},n}$: the best joint score (geometry + appearance) among all candidates for node~$v$, i.e.\ $\max_{c \in \mathcal{C}_v}\big(\!-\|\mathbf{p}_c - \hat{\mathbf{c}}_v(\hat\theta)\|^2 / (2\sigma_v^2) + \log w_{\text{app}}(v,c)\big)$.
\item $\log p_{\text{best\_sim},n}$: the appearance-only ceiling---the best $\log w_{\text{app}}(n,c)$ among the winning candidate, representing the score if geometry were perfect.
\item $\log p_{\text{miss},v} = \log p_{\text{miss}}$: the missing-node baseline.
\end{itemize}
The per-node confidence normalises the match quality relative to the best achievable:
\begin{equation}
c_v = \text{clip}\!\left(\frac{\log p_{\text{match},v} - \log p_{\text{miss},v}}{\log p_{\text{best\_sim},v} - \log p_{\text{miss},v}},\; 0,\; 1\right).
\end{equation}
When $\log p_{\text{match},v} \geq \log p_{\text{best\_sim},v}$ (near-perfect geometry), $c_v \to 1$. When $\log p_{\text{match},v} \leq \log p_{\text{miss},v}$ (worse than the missing baseline), $c_v = 0$. When the best candidate's similarity itself falls below the missing baseline ($\log p_{\text{best\_sim},v} < \log p_{\text{miss},v}$), the denominator is non-positive and $c_v$ is set to $0$---matching to a very weak candidate should not contribute positively.

\paragraph{Weighted aggregation.}
The overall likelihood confidence is a locality-weighted average:
\begin{equation}
\tilde{p}(Z\mid \bm{\theta}) = \frac{\sum_{v\in V_\mathrm{demo}} w_{\text{loc}}(v)\, c_v}{\sum_{v\in V_\mathrm{demo}} w_{\text{loc}}(v)}.
\end{equation}
This gives more influence to nearby nodes whose spatial predictions are more reliable. A high $\tilde{p}(Z\mid \bm{\theta})$ indicates that the predicted position is geometrically consistent with multiple context node observations; a low value signals that nodes cannot find their expected matches near the prediction, often because the UI layout has changed substantially.

\subsubsection{Spatial Confidence}\label{sec:appendix_spatial_conf}

Spatial confidence measures how tightly the particle population agrees on the target location, independent of the context-candidate likelihood.

\paragraph{Posterior predictive variance.}
After SMC convergence, we have $N$ weighted particles $\{(\theta^{(j)}, w^{(j)})\}$ with positions $\mathbf{x}^{(j)} = (\theta^{(j)}_x, \theta^{(j)}_y)$. We compute the weighted mean and covariance:
\begin{equation}
\boldsymbol{\mu} = \sum_j w^{(j)} \mathbf{x}^{(j)}, \qquad
\Sigma = \sum_j w^{(j)} (\mathbf{x}^{(j)} - \boldsymbol{\mu})(\mathbf{x}^{(j)} - \boldsymbol{\mu})^\top.
\end{equation}
Under an isotropic Gaussian approximation with variance $\bar\sigma^2 = \tfrac{1}{2}\operatorname{tr}(\Sigma)$, the probability that the true position lies within radius~$r$ of the mean is:
\begin{equation}\label{eq:spatial_conf}
C_{\text{spatial}} = 1 - \exp\!\left(-\frac{r^2}{2\,\bar\sigma^2}\right).
\end{equation}
This is the CDF of a Rayleigh distribution, which arises as the distribution of distances from the mean of a 2-D isotropic Gaussian.

\paragraph{Density-adaptive radius.}
The acceptance radius adapts to context node density:
\begin{equation}
r = r_{\text{base}} + \alpha \cdot \sigma_{\text{loc}},
\end{equation}
where $r_{\text{base}} = 50$\,px and $\alpha = 0.2$. In sparse UIs ($\sigma_{\text{loc}}$ large), fewer nodes provide weaker constraints, so particles naturally spread more. A fixed radius would over-penalise this expected spread and produce artificially low confidence. Scaling $r$ by $\sigma_{\text{loc}}$ ensures that the acceptance region grows with the spatial scale of the problem.

\paragraph{Behaviour.}
\begin{itemize}[nosep]
\item \emph{Dense UI} ($\sigma_{\text{loc}} \approx 40$\,px, $r \approx 58$\,px): nearly the base radius; tight convergence required.
\item \emph{Moderate UI} ($\sigma_{\text{loc}} \approx 200$\,px, $r \approx 90$\,px): tolerates wider particle spread.
\item \emph{Sparse UI} ($\sigma_{\text{loc}} \approx 500$\,px, $r \approx 150$\,px): substantial tolerance for the limited evidence.
\item \emph{Tight convergence} ($\bar\sigma \approx 5$\,px, $r = 50$): $C_{\text{spatial}} \approx 1.0$.
\item \emph{Wide scatter} ($\bar\sigma \approx 100$\,px, $r = 50$): $C_{\text{spatial}} \approx 0.12$.
\item \emph{Degenerate} (all particles at same point): $\operatorname{tr}(\Sigma) = 0 \Rightarrow C_{\text{spatial}} = 1$.
\end{itemize}

\paragraph{Why multiplicative combination.}
The product $C = \tilde{p}(Z\mid \bm{\theta}) \times C_{\text{spatial}}$ requires \emph{both} factors to be high for the final confidence to be high:
\begin{itemize}[nosep]
\item High $\tilde{p}(Z\mid \bm{\theta})$, low $C_{\text{spatial}}$: nodes match well individually, but particles are scattered across multiple modes $\Rightarrow$ ambiguous posterior, low confidence (correct: the prediction is unreliable).
\item Low $\tilde{p}(Z\mid \bm{\theta})$, high $C_{\text{spatial}}$: particles converge tightly, but to a position where nodes do not match $\Rightarrow$ confidently wrong, low confidence (correct: the prediction is likely incorrect).
\item High $\tilde{p}(Z\mid \bm{\theta})$, high $C_{\text{spatial}}$: strong context evidence \emph{and} tight particle agreement $\Rightarrow$ high confidence.
\end{itemize}
This decomposition separates two failure modes that the readiness checker must detect: observation mismatch (context confidence) and posterior uncertainty (spatial confidence).

\subsection{SMC Algorithm}\label{sec:appendix_smc_algorithm}

Rather than directly sampling the often sharply peaked and multi-modal posterior, the SMC sampler uses likelihood tempering to define a sequence of intermediate targets:
\begin{equation}
\pi_{\beta}(\theta) \propto p(Z\mid\theta)^{\beta}\,p(\theta),
\qquad 0=\beta_0 < \beta_1 < \cdots < \beta_S = 1.
\end{equation}
Intuitively, $\beta$ ``fades in'' the likelihood: $\pi_{\beta=0}$ is the prior and $\pi_{\beta=1}$ is the true posterior. We choose $\beta_{s+1}$ adaptively to maintain a desired effective sample size (ESS) under the incremental importance weights, with an optional cap on $\Delta\beta$ to ensure gradual annealing.

At each stage, we apply importance reweighting, resampling, and a Markov chain Monte Carlo (MCMC) rejuvenation step targeting $\pi_{\beta_s}$. This combination prevents particle degeneracy and improves mixing between modes. Concretely:
\begin{algorithm}[h]
\caption{Tempered SMC Sampler for Static UI Localization}
\label{alg:smc}
\begin{algorithmic}[1]
\REQUIRE Context-to-target displacement vectors $\{\mathbf{r}_i\}$ and candidate sets $\{\mathcal{C}_i\}$
\ENSURE Predicted click position $\hat{\mathbf{t}}$ and confidence
\STATE Initialize particles $\{\theta^{(j)}\}_{j=1}^{N}$ from proposal (fallback: uniform over screen bounds)
\STATE Set $\beta_0 \leftarrow 0$
\FOR{$s = 0$ to $S-1$}
    \STATE Choose $\beta_{s+1}\in(\beta_s,1]$ to meet an ESS target (optionally cap $\Delta\beta$)
    \STATE Incremental reweight: $\tilde{w}^{(j)} \propto w^{(j)} \cdot p(Z\mid\theta^{(j)})^{(\beta_{s+1}-\beta_s)}$
    \STATE Normalize: $w^{(j)} \leftarrow \tilde{w}^{(j)} / \sum_k \tilde{w}^{(k)}$
    \STATE Resample: draw $\{\theta^{(j)}\}$ with replacement according to $w$ (optionally ESS-gated)
    \STATE Rejuvenate (MCMC): for each particle, run $L$ MH steps targeting $\pi_{\beta_{s+1}}(\theta)\propto p(Z\mid\theta)^{\beta_{s+1}}p(\theta)$
    \STATE If confidence $> c_{\min}$ or ($\beta\!=\!1$ and confidence stable): break
\ENDFOR
\STATE Optional refinement: with $\beta=1$, repeat resample+MH for a few additional steps to tighten concentration
\STATE Output $\hat{\mathbf{t}}$ as the densest particle cluster mean (grid-based clustering)
\end{algorithmic}
\end{algorithm}

The Metropolis-Hasting (MH) proposal is a symmetric Gaussian random walk. Step sizes are scaled by the current particle spread (so particles contract instead of re-scattering near $\beta\!=\!1$) and modulated by an intensity factor. During tempering ($\beta<1$), intensity decays to progressively focus particles. After $\beta$ reaches~$1$, intensity resets with a floor so that MH retains enough step size to merge surviving sub-clusters against the full posterior. The loop may exit early when confidence exceeds a threshold or when $\beta\!=\!1$ and confidence has stabilized.

\section{UI Graph and Similarity Details}

\subsection{UI Graph Construction}\label{sec:appendix_ui_graph}

The UI graph is constructed from screenshots in three steps. First, we extract UI elements from the screenshot using computer vision detectors. We finetuned OmniParser for icon detection and OCR for text extraction. Each detected element becomes a node with its bounding box coordinates. Second, we compute visual features for each element using IconCLIP (ViT-B-32)~\cite{iconclip} to produce icon embeddings. Third, we build the graph structure by connecting spatially nearby elements. We use $k$-nearest neighbors based on element centers to create edges, typically with $k=5$. This creates a graph where edges represent spatial proximity rather than semantic relationships.

The resulting graph captures both the visual appearance and spatial layout of the UI. Elements that are close together in the interface are connected by edges. This structure allows the retrieval algorithm to use context from neighboring elements when matching across different UI states.

\subsection{Similarity Computation}\label{sec:appendix_similarity}

We compute similarity between nodes based on their type and content. For textual elements such as buttons with text or text fields, we use fuzzy string matching. The fuzzy matching uses the Levenshtein distance ratio (via RapidFuzz) with length-adaptive tolerance that provides more forgiveness for OCR errors in short strings, returning 1.0 for exact matches and smoothly degrading for partial matches. We then combine this text similarity with icon embedding similarity:
\begin{equation}
s(v_d, v_c) = 0.9 \cdot s_{\text{text}}(v_d, v_c) + 0.1 \cdot \cos(\mathbf{e}_d, \mathbf{e}_c),
\end{equation}
where $\mathbf{e}_d, \mathbf{e}_c$ are icon embeddings. For non-textual elements such as pure icons or images, we use only icon embedding similarity: $s(v_d, v_c) = \cos(\mathbf{e}_d, \mathbf{e}_c)$.

\subsection{Scale Prior and Geometric Tolerance}\label{sec:appendix_scale_prior}

The scale prior for each axis is a mixture of two log-normals:
\begin{equation}
p(s_x) = w\,\mathrm{LogNormal}(s_x;\,0,\,\sigma_{s,x}) + (1-w)\,\mathrm{LogNormal}(s_x;\,\mu_x,\,\sigma_{s,x}),
\end{equation}
and analogously for $s_y$ with mean $\mu_y = \log(H_{\text{live}}/H_{\text{demo}})$. The identity component ($\mu=0$, weight $w$) captures the case where no resizing occurs; the ratio component ($\mu_x = \log(W_{\text{live}}/W_{\text{demo}})$, weight $1-w$) captures a full proportional rescaling of the window. Using per-axis ratios instead of a single isotropic mean accounts for independent horizontal and vertical rescaling. Penalties are applied symmetrically in log-space so that shrink and expand are equally penalized. The $y$-axis prior uses a tighter $\sigma_{s,y}$ since vertical layout is typically more stable. Hard bounds are enforced on $(s_x,s_y)$.

The geometric tolerance is \emph{size-aware}:
\begin{equation}
\sigma_i = \sigma_{\text{base}} + \alpha\, d_i + \beta\, \min(w_i,h_i),
\end{equation}
where $d_i=\|\mathbf{c}_i^{\text{demo}}-\mathbf{t}_{\text{demo}}\|_2$ is the distance from the context node to the target in the demonstration and $(w_i,h_i)$ is the context node element size. This makes localization stricter for small UI elements and more tolerant for large ones, while also accounting for the fact that distant nodes have inherently less precise geometric constraints.

\section{Precheck Pipeline}\label{sec:appendix_precheck}

After executing an action, the environment typically needs time to settle (e.g., waiting for a page to load or an animation to finish). The precheck pipeline exploits this idle time by speculatively processing upcoming steps in a background thread, so that when the runner advances to the next step, a cached result may already be available.

The pipeline operates as follows. When the runner finishes step~$N$, it submits the current observation to the pipeline and dispatches the action to the environment. While the environment executes, the pipeline processes step~$N\!+\!1$ (and optionally step~$N\!+\!2$) using the last available observation. When the environment returns a new observation, the runner calls \texttt{collect()} to retrieve any completed results. If the precheck result for step~$N\!+\!1$ has sufficiently high confidence, the runner uses it directly, skipping redundant parsing and retrieval. Otherwise, the cached result is discarded and the step is processed normally with the fresh observation.

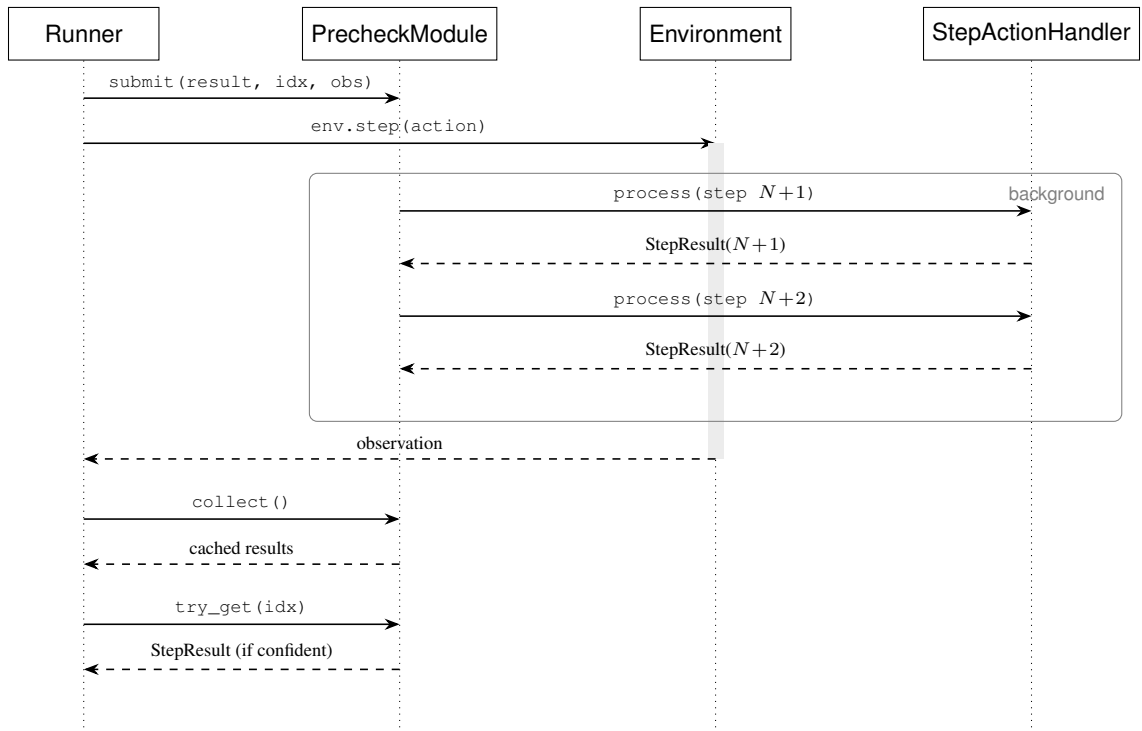
\begin{figure*}[h]
\centering
\begin{tikzpicture}[
    >=Stealth,
    participant/.style={rectangle, draw, minimum height=0.7cm, minimum width=2cm, font=\footnotesize\sffamily},
    msg/.style={->, semithick},
    rmsg/.style={->, semithick, dashed},
]
    \node[participant] (runner) at (0,0) {Runner};
    \node[participant] (pipeline) at (4.2,0) {PrecheckModule};
    \node[participant] (env) at (8.4,0) {Environment};
    \node[participant] (handler) at (12.6,0) {StepActionHandler};

    \foreach \n in {runner, pipeline, env, handler} {
        \draw[dotted, thin] (\n.south) -- ++(0,-9.0cm);
    }

    \draw[msg] ($(runner.south)+(0,-0.5)$) -- node[above, font=\scriptsize] {\texttt{submit(result, idx, obs)}} ($(pipeline.south)+(0,-0.5)$);
    \draw[msg] ($(runner.south)+(0,-1.1)$) -- node[above, font=\scriptsize] {\texttt{env.step(action)}} ($(env.south)+(0,-1.1)$);

    \fill[gray!15] ($(env.south)+(-0.1,-1.1)$) rectangle ($(env.south)+(0.1,-5.3)$);

    \draw[rounded corners=3pt, gray, thin]
        ($(pipeline.south)+(-1.2,-1.5)$) rectangle ($(handler.south)+(1.2,-4.8)$);
    \node[font=\scriptsize\sffamily, gray, anchor=north east] at ($(handler.south)+(1.1,-1.55)$) {background};

    \draw[msg] ($(pipeline.south)+(0,-2.0)$) -- node[above, font=\scriptsize] {\texttt{process(step $N\!+\!1$)}} ($(handler.south)+(0,-2.0)$);
    \draw[rmsg] ($(handler.south)+(0,-2.7)$) -- node[above, font=\scriptsize] {StepResult($N\!+\!1$)} ($(pipeline.south)+(0,-2.7)$);

    \draw[msg] ($(pipeline.south)+(0,-3.4)$) -- node[above, font=\scriptsize] {\texttt{process(step $N\!+\!2$)}} ($(handler.south)+(0,-3.4)$);
    \draw[rmsg] ($(handler.south)+(0,-4.1)$) -- node[above, font=\scriptsize] {StepResult($N\!+\!2$)} ($(pipeline.south)+(0,-4.1)$);

    \draw[rmsg] ($(env.south)+(0,-5.3)$) -- node[above, font=\scriptsize] {observation} ($(runner.south)+(0,-5.3)$);

    \draw[msg] ($(runner.south)+(0,-6.1)$) -- node[above, font=\scriptsize] {\texttt{collect()}} ($(pipeline.south)+(0,-6.1)$);
    \draw[rmsg] ($(pipeline.south)+(0,-6.7)$) -- node[above, font=\scriptsize] {cached results} ($(runner.south)+(0,-6.7)$);

    \draw[msg] ($(runner.south)+(0,-7.5)$) -- node[above, font=\scriptsize] {\texttt{try\_get(idx)}} ($(pipeline.south)+(0,-7.5)$);
    \draw[rmsg] ($(pipeline.south)+(0,-8.1)$) -- node[above, font=\scriptsize] {StepResult (if confident)} ($(runner.south)+(0,-8.1)$);

\end{tikzpicture}
\caption{Precheck pipeline sequence diagram. While the environment executes the current action (gray bar), the pipeline speculatively processes upcoming steps in a background thread. The runner retrieves cached results via \texttt{try\_get}, using them directly if confidence is sufficient.}
\label{fig:precheck_pipeline}
\end{figure*}

\section{Workflow Template Example}\label{sec:appendix_workflow_example}

Below is an example workflow template generated from a single demonstration of drafting an email. The template contains metadata (name, title, description), parameterisable variables with placeholder defaults, and an ordered list of steps with natural-language action descriptions. Variables such as \texttt{\{\{recipient\_email\}\}} are automatically extracted during the build phase and substituted at execution time.

\begin{lstlisting}[language=yaml,title={\footnotesize\texttt{workflow.yaml}}]
workflow_id: '20260313_110944_3'
workflow_name: draft_email
workflow_title: Draft Email
description: Compose and save a draft email with recipient address, subject line,
  and message content
running_config:
  variable_values:
    recipient_email: email address
    subject: email subject
    email_content: email content
  injected_steps: []
category: ''
steps:
- step_number: 1
  Action: Click on Mail icon
  id: c2c927c9-2056-4935-b225-93275e9e655d
- step_number: 2
  Action: Click on Compose button
  id: 0559d739-6660-4ec6-b4c7-1493cac585bf
- step_number: 3
  Action: 'Type text: {{recipient_email}}'
  id: dddef290-3c3f-42a3-a266-5730582fe6f8
- step_number: 4
  Action: 'Press key: tab'
  id: 88f6250f-3f02-44e4-ac0b-099174fac6e8
- step_number: 5
  Action: 'Press key: tab'
  id: d6174e85-e256-4deb-bc12-f654d151b722
- step_number: 6
  Action: 'Type text: {{subject}}'
  id: d18f4cf2-1dd8-4a04-81df-9b8124880f57
- step_number: 7
  Action: 'Press key: tab'
  id: 8e6de124-c15e-45f3-bac7-1de89e0a1547
- step_number: 8
  Action: 'Type text: {{email_content}}'
  id: 30c9f2a6-54fb-4827-b197-7d82a2e65517
- step_number: 9
  Action: 'Press hotkey: cmd + s'
  id: b0c473e9-393e-47c7-93b5-da8adabcc55b
- step_number: 10
  Action: Click on close button
  id: bd0e41cc-216d-4ea4-88b0-aecf7c390121
\end{lstlisting}

The accompanying metadata file stores build-time configuration, extracted variable descriptions, and references to the step data used during execution:

\begin{lstlisting}[language=json,title={\footnotesize\texttt{metadata.json}}]
{
  "format_version": "2.0",
  "workflow_id": "20260313_110944_3",
  "config_file": "storage/workflows/.../workflow.yaml",
  "data_file": "steps_data.json",
  "step_count": 10,
  "created_at": "2026-03-18T19:59:42.605361",
  "workflow_metadata": {
    "variables": [
      {
        "name": "recipient_email",
        "default_value": "email address",
        "description": "Email address of the recipient"
      },
      {
        "name": "subject",
        "default_value": "email subject",
        "description": "Subject line of the email"
      },
      {
        "name": "email_content",
        "default_value": "email content",
        "description": "Body content of the email message"
      }
    ],
    "metadata": {
      "scale_factor": 2.0,
      "builder_config": {
        "ui_parser_config": {
          "text_feature": "sentence_e5",
          "vision_feature": "iconclip",
          "ui_extractors": [
            "batch_omniparser",
            "ocrmac_detector"
          ]
        },
        "knn_k": 8
      }
    }
  }
}
\end{lstlisting}

The referenced \texttt{steps\_data.json} is keyed by step UUID and contains the per-step data used during execution. Its structure (with embeddings and lists abbreviated) is shown below:

\begin{lstlisting}[language=json,title={\footnotesize\texttt{steps\_data.json} (abbreviated)}]
{
  "<step_uuid>": {
    "step_id": "<uuid>",
    "step_number": 1,
    "action_type": "click",
    "description": "Click on Mail icon",
    "step_subgraph": {
      "target_element_id": 7,
      "image_size": [W, H],
      "click_coordinates": [x, y],
      "ui_graph": {
        "G": {
          "directed": false,
          "nodes": [
            {
              "pos": [x, y, w, h],
              "icon_emb": [/* 512-d vector */],
              "text_emb": [/* 384-d vector or null */],
              "attrs": {"type": "icon|text", "content": "..."},
              "id": 0
            }
            /* ... more nodes ... */
          ],
          "links": [
            {"source": 0, "target": 10}
            /* ... kNN edges ... */
          ]
        }
      },
      "window_bounds": [x, y, w, h],
      "knn_k": 8,
      "scale_factor": 1.0,
      "offset_x": -9.0,
      "offset_y": -2.59
    },
    "variables": [],
    "values": {},
    "pause_duration": 1.73,
    "active_app_name": "Dock"
  }
  /* ... one entry per step ... */
}
\end{lstlisting}